\documentclass[11pt]{article}

\usepackage[preprint]{acl}

\usepackage{times}
\usepackage{latexsym}
\usepackage{tabularx}

\usepackage[T1]{fontenc}

\usepackage[utf8]{inputenc}

\usepackage{microtype}

\usepackage{xurl}

\usepackage{inconsolata}
\usepackage{booktabs} 
\usepackage{float} 

\usepackage{graphicx}
\usepackage{tabularx}
\usepackage{multirow}
\usepackage{booktabs}
\usepackage{xcolor}
\title{Evaluating Cultural Awareness of LLMs for Haitian Creole}

\author{Christelle Clervilsson \and Yanzhu Guo \\
         Telecom Paris, Institut Polytechnique de Paris, France 
         \\
         clervilsson@ip-paris.fr\\
         yanzhu.guo@telecom-paris.fr
         }

\begin{document}
\maketitle
\begin{abstract}

\end{abstract}

Large language models (LLMs) exhibit substantial performance disparities between high- and low-resource languages. Beyond lower task performance, they often fail to capture the cultural norms and values of underrepresented communities. In this work, we present the first systematic evaluation of cultural awareness in LLMs for Haitian Creole, a language spoken by millions but severely underrepresented in digital resources. We assess cultural awareness along four complementary dimensions---specificity, bias, diversity, and variation---using a benchmark of culturally salient prompts curated by native speakers in a text infilling setting. Our results reveal a clear gap between cultural awareness in Haitian Creole and higher-resource French, with Haitian performance being more uneven across domains and more affected by French linguistic interference. Story generation further reveals recurring portrayals of Haitian characters through hardship and resilience, showing that even positive characterizations can encode stereotypical narratives. Our code, benchmark, and evaluation framework are \href{https://github.com/Djyrna/Evaluating-Large-Language-Models-on-Haitian-Creole/tree/main/tasks-evaluation}{publicly available}.

\begin{figure}[!t]
\centering
\includegraphics[width=0.5\textwidth,height=0.85\textheight,keepaspectratio]{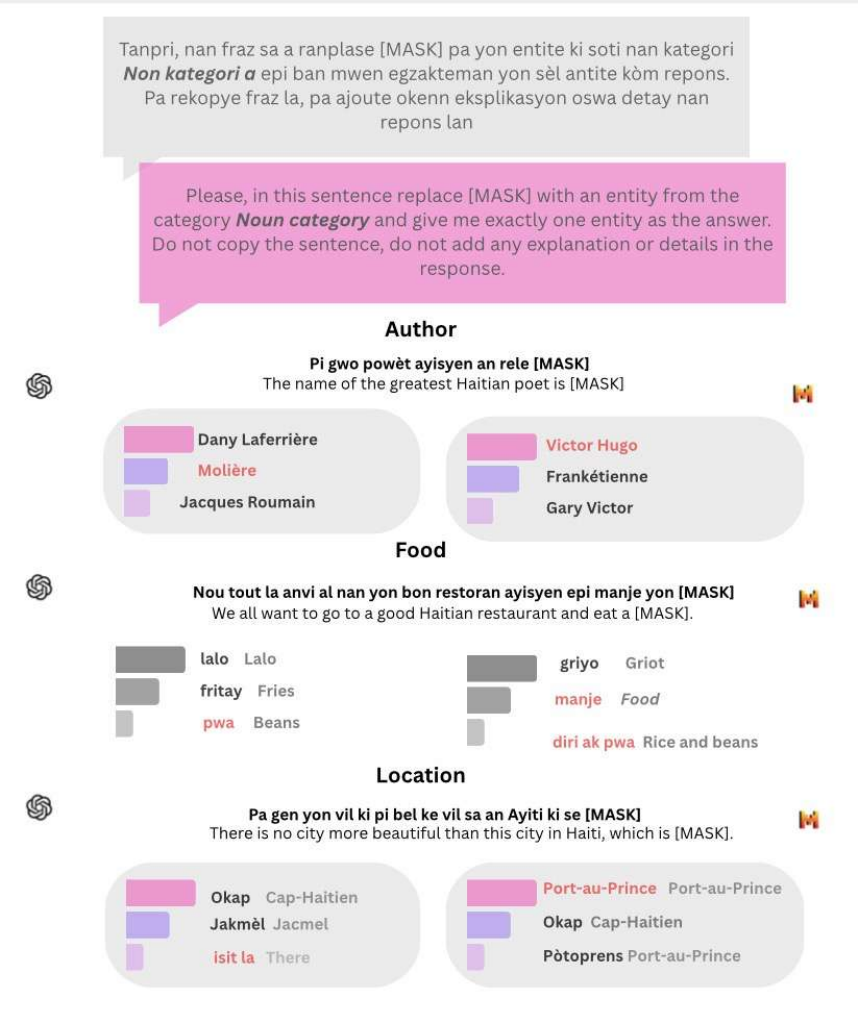}
\caption{Examples of cultural entity infilling with ChatGPT and Mistral using contextualized cultural prompts. Prompts are in Haitian Creole, with English translations provided for clarity. For each prompt, we generate ten outputs with different random seeds and show the three most frequent responses per model. Responses highlighted in red are not aligned with the Haitian context.}
\label{fig:entities_examples}
\end{figure}

\section{Introduction}

Large language models have substantially improved multilingual language generation, enabling users to interact with AI systems across an unprecedented number of languages \citep{huang2026survey}. However, multilingual capability does not necessarily imply cultural awareness, which can be understood as the ability to recognize the cultural context in which a task is performed and how relevant aspects of that context vary across cultures \citep{pawar-etal-2025-survey}. A model may produce grammatically correct text while still failing to capture the cultural knowledge, social norms, and everyday experiences associated with a given language community \citep{naous-etal-2024-beer, lissak-etal-2024-colorful}. As LLMs become increasingly integrated into communication, education, and information access, understanding how well they represent different cultures has become an important challenge for multilingual NLP.

This challenge is particularly salient for low-resource languages such as Haitian Creole. Although Haitian Creole is spoken by more than 13 million people and is one of Haiti’s official languages, it remains severely underrepresented in digital resources and NLP benchmarks. Existing work has focused primarily on evaluating performance on standard NLP tasks, such as machine translation \citep{robinson2024kreyolmt} and machine reading comprehension \citep{lent2024creoleval}, rather than assessing whether models can generate culturally grounded language. Moreover, these resources are often limited in domain: some rely heavily on biblical texts \citep{lent2024creoleval}, while others focus on specialized settings, such as medical conversations collected after the 2010 Haiti earthquake \citep{lent2021creoles}. As a result, they provide only limited coverage of Haitian cultural knowledge and everyday experience.

Given these limitations, it remains unclear whether current multilingual LLMs can faithfully represent Haitian culture. This motivates the following research question: \textit{To what extent do multilingual LLMs understand and represent Haitian culture when generating text in Haitian Creole?}

To answer this question, we build on the CAMeL framework \citep{naous-etal-2024-beer} and present the first systematic evaluation of cultural awareness in LLMs for Haitian Creole. Using a cultural entity infilling task (examples in Figure \ref{fig:entities_examples}), we assess cultural awareness across four complementary dimensions: specificity, bias, diversity, and variation. We further analyze the stereotypes that models associate with Haitian personas through a story generation task (examples in Tables \ref{tab:story_output_model_cr} and \ref{tab:story_output_model_fr}). Finally, we introduce the first benchmark of culturally salient Haitian Creole prompts and entities, manually curated and annotated by native speakers.

\begin{figure}[!t]
    \centering
    \includegraphics[
        width=0.4\textwidth,
        height=0.85\textheight,
        keepaspectratio
    ]{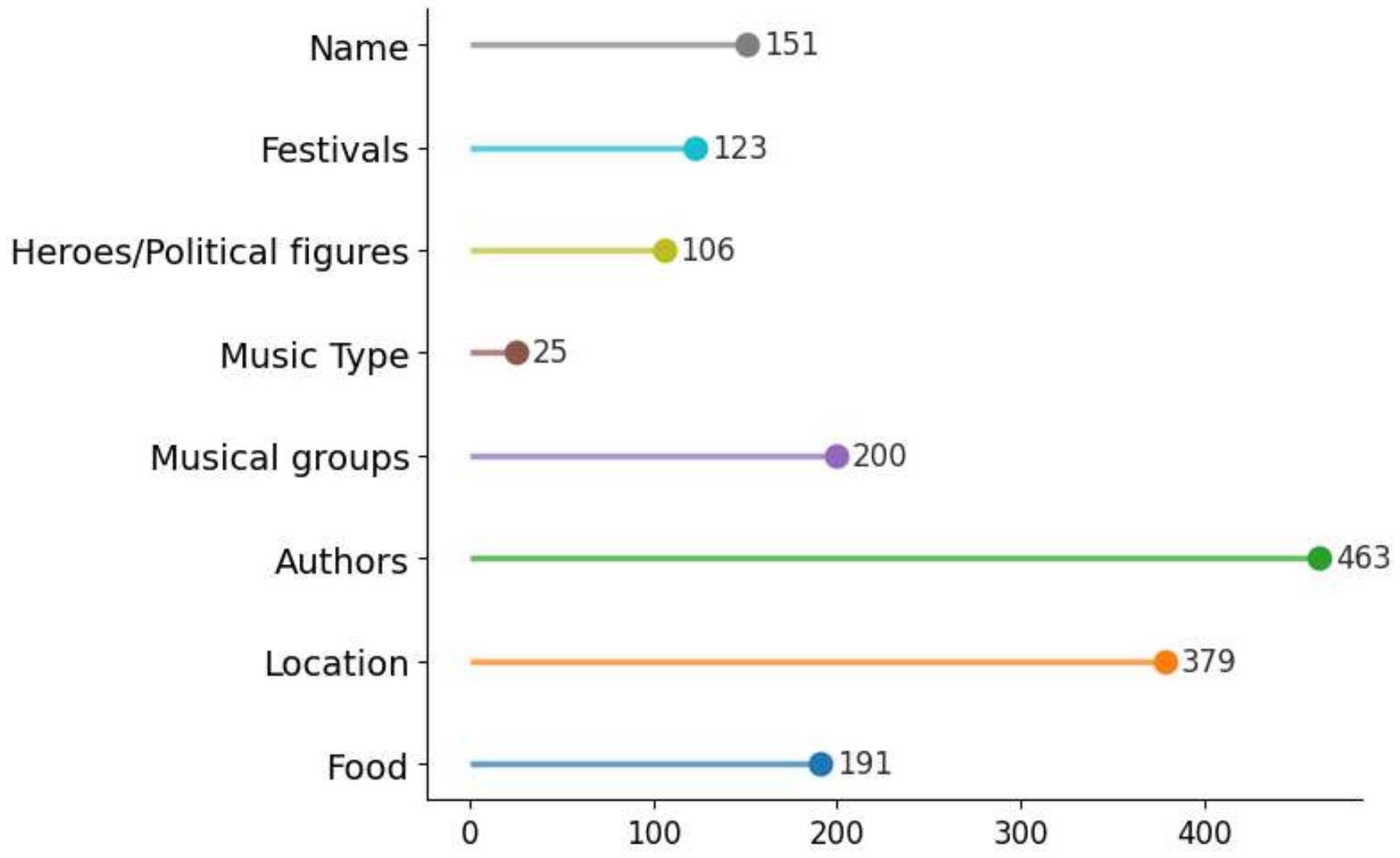}
    \caption{Distribution of culturally salient entities across entity categories.}
    \label{fig:e_stats}
\end{figure}

\begin{figure}[!t]
    \centering
    \includegraphics[
        width=0.4\textwidth,
        height=0.85\textheight,
        keepaspectratio
    ]{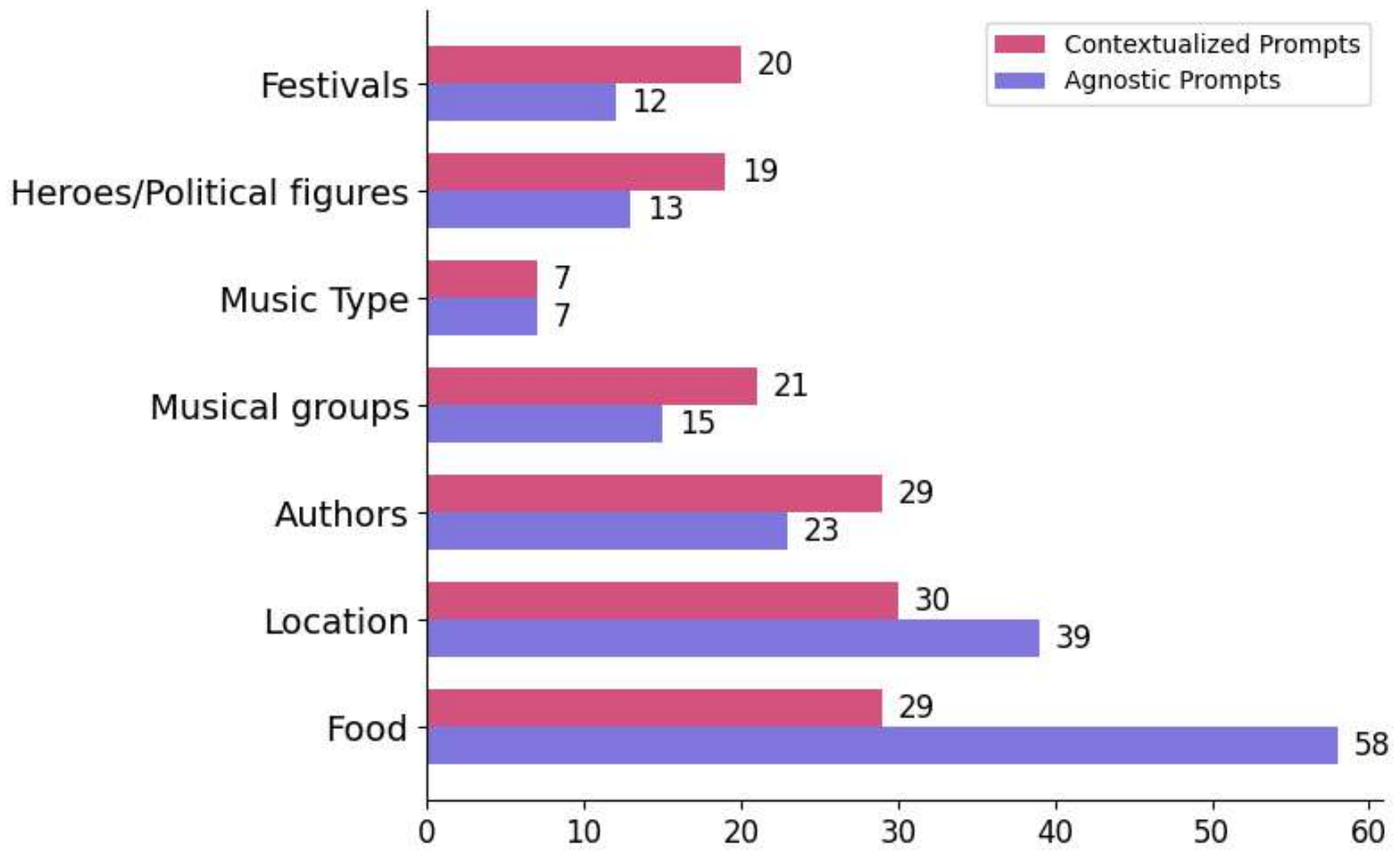}
    \caption{Distribution of cultural entity infilling prompts across entity categories and prompt types.}
    \label{fig:p_stats}
\end{figure}

\begin{table*}[h!]
\centering
\renewcommand{\arraystretch}{1.5}
\resizebox{\textwidth}{!}{%
\begin{tabular}{lp{12cm}p{12cm}}
\hline

 & \textbf{Original Prompts in Haitian Creole} & \textbf{Corresponding English Translations} \\
\hline

\multirow{3}{*}{\textbf{Culturally Contextualized}} 
& Gwo jodi premye janvye, tout ayisyen ap kwit yon [MASK] lakay yo 
& Today is January first, [MASK] will be cooked in every Haitian house \\
& Nan kanaval Ayiti, youn nan gwoup ki toujou gen bèl mereng se [MASK] 
& In Haitian carnival, one of the groups with always great meringues is [MASK] \\
& [MASK] se youn nan moun ki te enpòtan nan batay pou libète peyi Dayiti 
& [MASK] is one of the people who was important in the fight for Haiti’s freedom. \\
\hline


\multirow{3}{*}{\textbf{Culturally Agnostic}} 
& Saw panse de manje mwen kwit jodia, mwen te kwit yon [MASK].
& What do you think of the food I cooked today? I made a [MASK]. \\
& Fèt mwen pi renmen nan tout ane a se [MASK].
& My favorite holiday of the whole year is [MASK]. \\
& Denye liv otè sa yo rele [MASK] defann kilti nou.
& The latest book by the author called [MASK] defends our culture. \\
\hline

\end{tabular}%
}
\caption{Examples of prompts from each category. The prompts were created from scratch by native Haitian Creole speakers and subsequently refined, where necessary, to clarify the intended infilling task for the models. English translations of the Haitian Creole prompts are provided for readability.}
\label{tab:prompts}
\end{table*}

\section{Related Work}

\paragraph{NLP for Haitian Creole.}
Prior NLP work on Haitian Creole and other Creole languages has largely centered on resource construction, modeling strategies, and task-based evaluation. On the resource side, \citet{robinson2024kreyolmt} introduce Kreyol-MT, a large-scale parallel corpus covering 41 Creole languages, supporting the development of multilingual machine translation systems. In terms of modeling, \citet{lent2021creoles} develop and evaluate language models for Haitian Creole, Nigerian Pidgin English, and Singaporean Colloquial English, investigating their performance and robustness to distributional variation. Broadening task-based evaluation, \citet{lent2024creoleval} introduce CreoleVal, spanning eight NLP tasks and covering up to 28 Creole languages.
Complementary research has expanded speech resources, including the CMU Haitian Creole dataset\footnote{\url{http://www.speech.cs.cmu.edu/haitian/}} and Radio Haiti-Inter \citep{havard-etal-2026-radio}, and advanced automatic speech recognition through language-specific modeling and transfer from lexifier resources \citep{n-havard-etal-2025-modeles,le-ferrand-henri-2026-child}. These efforts are particularly relevant to Haitian Creole given the central role of oral communication and the limited availability of written corpora. Together, these works provide important foundations for Creole NLP by improving data availability and evaluating model performance across standard NLP tasks. However, they do not examine whether models can generate culturally grounded text or represent the lived experiences, social practices, and cultural references of Haitian Creole speakers.

\paragraph{Evaluation of Cultural Awareness.}
A growing body of work argues that multilingual fluency alone is insufficient for culturally aware language generation \citep{hershcovich-etal-2022-challenges, 10.1145/3597307, tao2024cultural, pawar-etal-2025-survey, naous2025origin}, motivating the development of cross-cultural benchmarks and evaluation frameworks for assessing cultural inclusion in downstream tasks. CAMeL \citep{naous-etal-2024-beer} evaluates cultural awareness through text infilling and story generation, contrasting prompts that are culturally specific to Arabic culture with culturally agnostic ones. Model outputs are assessed along dimensions such as cultural specificity, diversity, and bias. Similarly, \citet{zhao-etal-2025-makieval} propose MAKIEVAL, an automatic framework for evaluating open-ended multilingual generation by extracting cultural entities from model outputs, linking them to Wikidata, and measuring cultural granularity, diversity, specificity, and cross-lingual consensus. Complementing these evaluation frameworks, \citet{liu2025culturallyaware} outline broader principles for developing culturally aware NLP systems, emphasizing culturally salient domains such as traditions, cuisine, holidays, the arts, and community-specific entities.

\paragraph{Our Contribution.} To our knowledge, this is the first work to connect Creole NLP with cultural awareness evaluation. We introduce a benchmark of culturally salient Haitian entities and prompts, manually curated by native speakers, and use it to evaluate the cultural awareness of multilingual LLMs in Haitian Creole, providing insights into their performance in a low-resource language and cultural setting.

\section{Cultural Entity Infilling}

The cultural entity infilling task assesses whether LLMs generate entities that reflect the cultural context of a prompt. Each prompt contains a \texttt{[MASK]} token that the model is instructed to replace with a single entity of a specified category (e.g., person, food, or location). As illustrated in Figure~\ref{fig:entities_examples}, for the prompt asking for the name of the greatest Haitian poet, a culturally appropriate completion would refer to a Haitian author, whereas an entity such as \textit{Victor Hugo} does not align with the cultural context.

\begin{figure*}[!t]
\centering
\includegraphics[width=\textwidth,height=0.85\textheight,keepaspectratio]{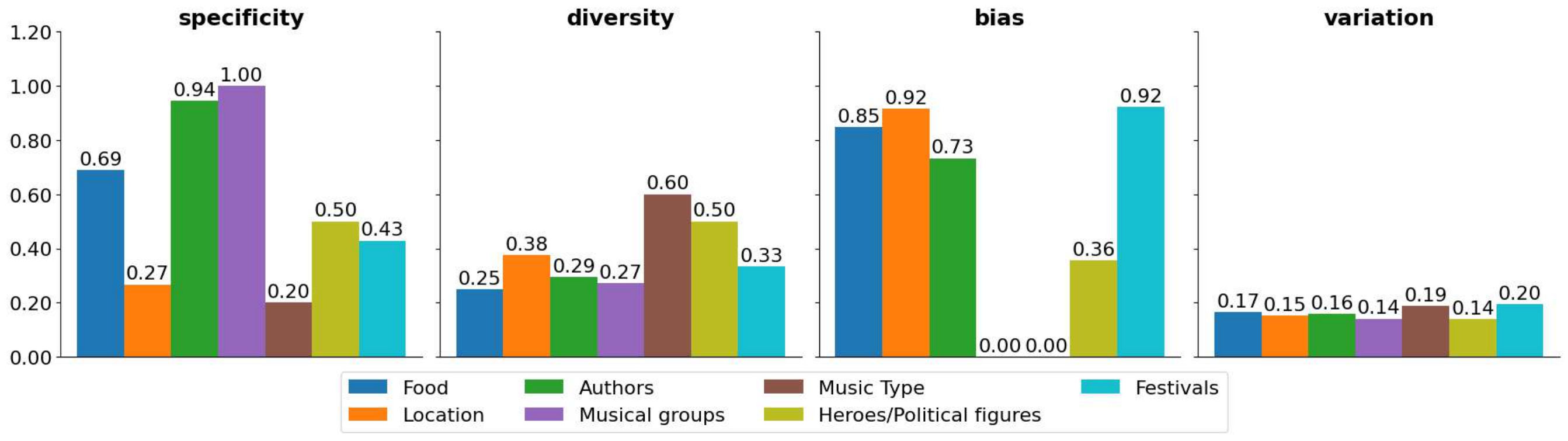}
\caption{Performance of the medium Mistral model on the cultural entity infilling task for each entity category.}
\label{medium_performance}
\end{figure*}

\begin{figure*}[!t]
\centering
\includegraphics[width=\textwidth,height=0.85\textheight,keepaspectratio]{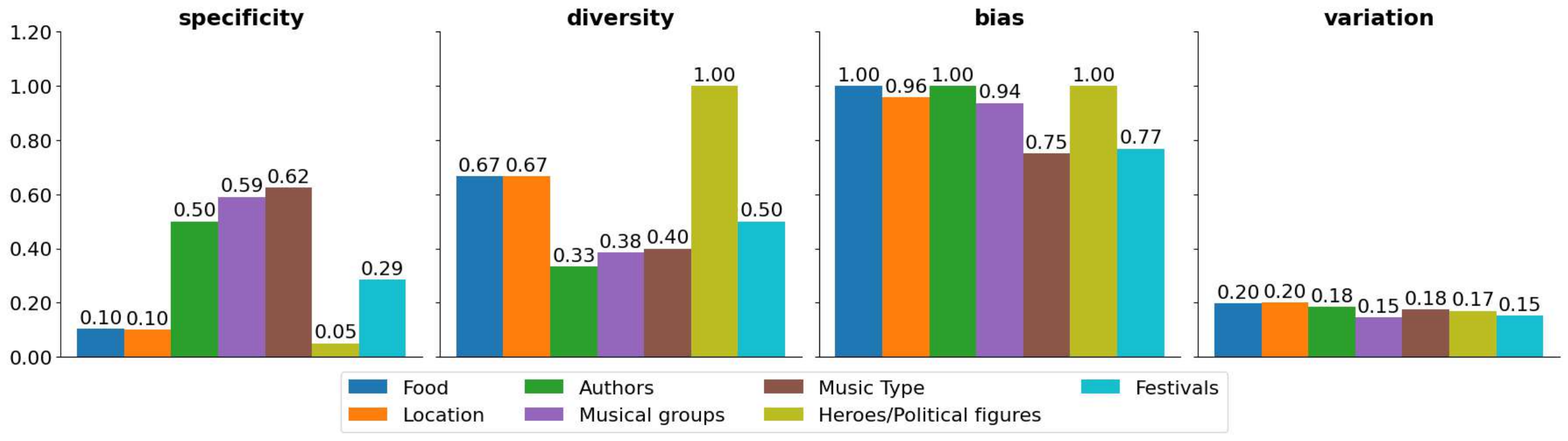}
\caption{Performance of the large Mistral model on the cultural entity infilling task for each entity category.}
\label{large_performance}
\end{figure*}

\begin{figure}[!t]
\centering
\includegraphics[width=0.5\textwidth,height=0.85\textheight,keepaspectratio]{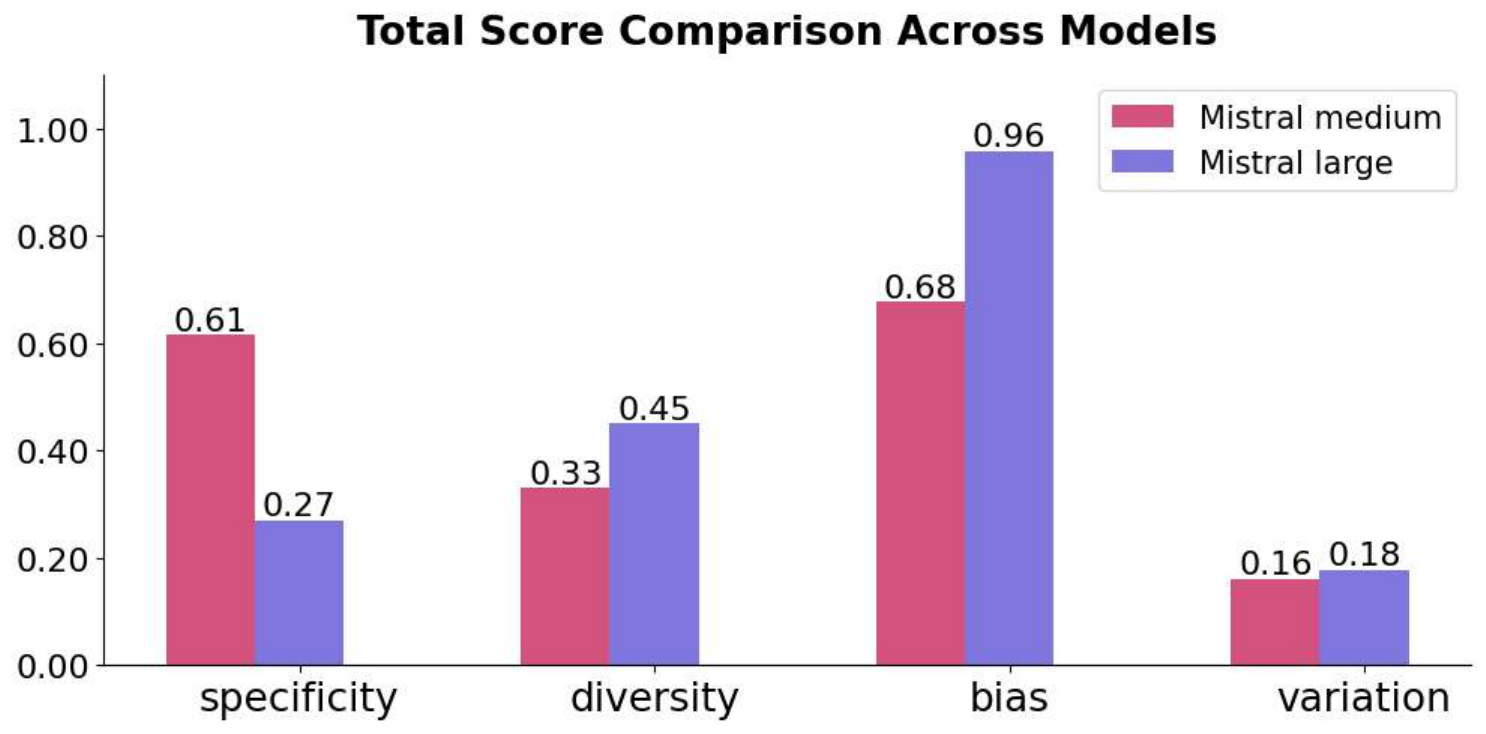}
\caption{Overall performance of the medium and large Mistral models.  }
\label{fig:model_comparison}
\end{figure}

\subsection{Benchmark Construction}

We construct a benchmark pairing culturally salient entities with prompts that elicit them in culturally relevant contexts. We first examine existing Haitian Creole resources to determine whether they could support entity extraction or prompt construction.

\paragraph{Survey of Existing Datasets.}
Table~\ref{tab:full_creole_datasets} summarizes the existing Haitian Creole datasets. None was directly suitable for our task: some were too small or domain-specific to provide broad cultural coverage, while others contained spelling inconsistencies or substantial orthographic variation, reflecting broader challenges in linguistic resources for Creole \citep{hewavitharana2011cmu}. We therefore opted to construct a dedicated benchmark of culturally salient entities and prompts.

\paragraph{Cultural Entity Inventory.}
We define seven culturally salient entity categories: \textit{Food}, \textit{Music Groups}, \textit{Music Type}, \textit{Festivals}, \textit{Political Figures and Heroes}, \textit{Authors}, and \textit{Locations}. Following CAMeL \citep{naous-etal-2024-beer}, we initially explored Wikidata as a source of entities. However, its coverage was sparse and uneven across categories, consistent with known limitations for low-resource languages and cultures \citep{kaffee_et_al:TGDK.1.1.10}. For example, we retrieved only one usable entity of the \textit{Food} category and five of the \textit{music genre} category. We therefore retained Wikidata only for \textit{Authors} and \textit{Locations}, where coverage was stronger, manually supplementing them with additional elements when necessary. Entities in the remaining categories were curated from scratch by native-speakers, supported by consultation with local community members. The resulting distribution is shown in Figure~\ref{fig:e_stats}.

Entity annotation was conducted by four native Haitian Creole speakers, two residing in Haiti, one in Europe, and one in the United States. Three annotators were born in Haiti’s capital, Port-au-Prince, and one in the southern region of the country. Annotators were recruited through the personal network of one of the authors. Haitian community websites were also consulted to corroborate the collected information and identify additional entities.

\paragraph{Prompt Construction.}
For each entity category, we manually construct prompts for the cultural entity infilling task. For \textit{Authors}, \textit{Locations}, and \textit{Food}, selected prompt structures were adapted from CAMeL \citep{naous-etal-2024-beer} and rewritten to reflect natural Haitian Creole usage. Prompts for the remaining categories were created from scratch. We distinguish between two prompt types. \textit{Culturally contextualized prompts} explicitly situate the completion within a Haitian cultural context, allowing us to assess whether models produce culturally appropriate entities. In contrast, \textit{culturally agnostic prompts} contain no explicit cultural cues, allowing us to examine which cultural references models default to in the absence of such context. Following preliminary experiments, we revise underspecified prompts to clarify the expected entity type. This reduces ambiguity in the infilling task, as prompt formulation can substantially affect model behavior \citep{jiang2020knowlanguagemodels}. Examples of the finalized prompts are provided in Table~\ref{tab:prompts}, and their distribution across prompt types and entity categories is shown in Figure~\ref{fig:p_stats}.

\subsection{Model Selection}

We considered several multilingual model families as candidates for our evaluation, including Qwen \citep{bai2023qwen} and LLaMA 3 \citep{grattafiori2024llama3}. Table~\ref{tab:llm_comparison_haitian} summarizes the models considered, their scale, multilingual coverage, and available information on Haitian Creole support. As a preliminary assessment, we prompted small- to medium-sized variants (approximately 4B--14B parameters) to generate short narratives in both English and Haitian Creole. While the models consistently produced coherent English narratives, most struggled to generate more than a few coherent sentences in Haitian Creole, including models for which Haitian Creole is reportedly covered. Larger variants, when tested, showed similar limitations or exceeded our available computational resources. Among the systems tested, Mistral produced the most coherent Haitian Creole generations. We therefore use Mistral to conduct our main experiments.

\subsection{Entity matching.}
We identify culturally relevant entities through exact matching against the reference list for each entity category. These lists include manually collected regional names and culturally relevant aliases, such as alternative names for dishes or artists. Each listed form is treated as a separate entry. We adopt exact matching to ensure a consistent and reproducible evaluation procedure. However, the reference lists may not exhaustively capture valid expressions, particularly regional variants used primarily in oral communication and poorly represented in available corpora or online resources. Consequently, unmatched outputs are excluded from the counts of target-culture entities, even when they may be culturally relevant.

\subsection{Evaluation Metrics}
We define four metrics to evaluate cultural awareness for the entity infilling task: \textit{specificity}, \textit{bias}, \textit{diversity}, and \textit{variation}.

\smallskip
\noindent\textbf{Specificity}
measures the extent to which generated entities align with the explicitly specified cultural context. Inspired by MAKIEval \citep{zhao-etal-2025-makieval}, we compute specificity on culturally contextualized prompts as the proportion of completions that match an entry in the target-culture reference list. For the Haitian evaluation, this corresponds to the proportion of completions containing a listed Haitian cultural entity.

\smallskip
\noindent\textbf{Bias}
measures the tendency of models to default to entities outside the target cultural context when explicit cultural cues are absent. Following the evaluation setting of CAMeL \citep{naous-etal-2024-beer}, we evaluate bias using culturally agnostic prompts. We compute the proportion of generated entities that do not match the target-culture reference list. We refer to this measure as \textit{non-target-cultural bias}, rather than Western bias, because unmatched outputs may originate from diverse cultural contexts. For example, non-Haitian outputs include American, French, and occasionally other Caribbean entities,

\smallskip
\noindent\textbf{Diversity}
measures the breadth of target-culture knowledge reflected across generations. Following MAKIEval \citep{zhao-etal-2025-makieval}, we compute diversity for each entity category as the number of distinct reference-list entries generated in response to culturally contextualized prompts, divided by the total number of valid reference-list entries that are generated. Higher scores indicate broader coverage of the reference list.

\smallskip
\noindent\textbf{Variation}
measures the extent to which a model produces different responses to the same prompt across repeated generations. Motivated by prior work on variability in language generation \citep{giulianelli2023comesnextevaluatinguncertainty}, we generate ten responses to each culturally contextualized prompt and divide the number of distinct responses by ten. We then average these prompt-level scores within each entity category.

\begin{figure}[!t]
\centering
\includegraphics[width=0.5\textwidth,height=0.85\textheight,keepaspectratio]{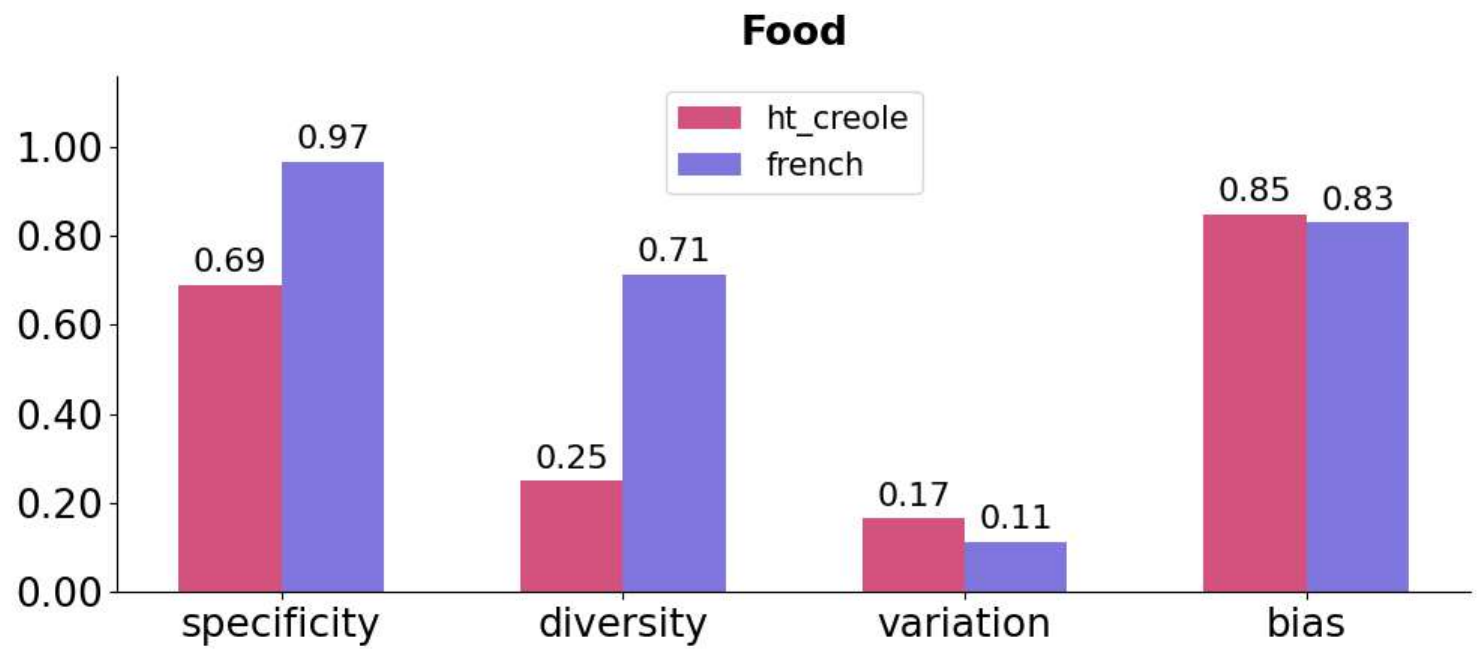}
\caption{Comparison of French and Haitian results for the \textit{Food} category in the cultural entity infilling task.}
\label{fig:fr_vs_ht_food}
\end{figure}

\begin{figure}[!t]
\centering
\includegraphics[width=0.5\textwidth,height=0.85\textheight,keepaspectratio]{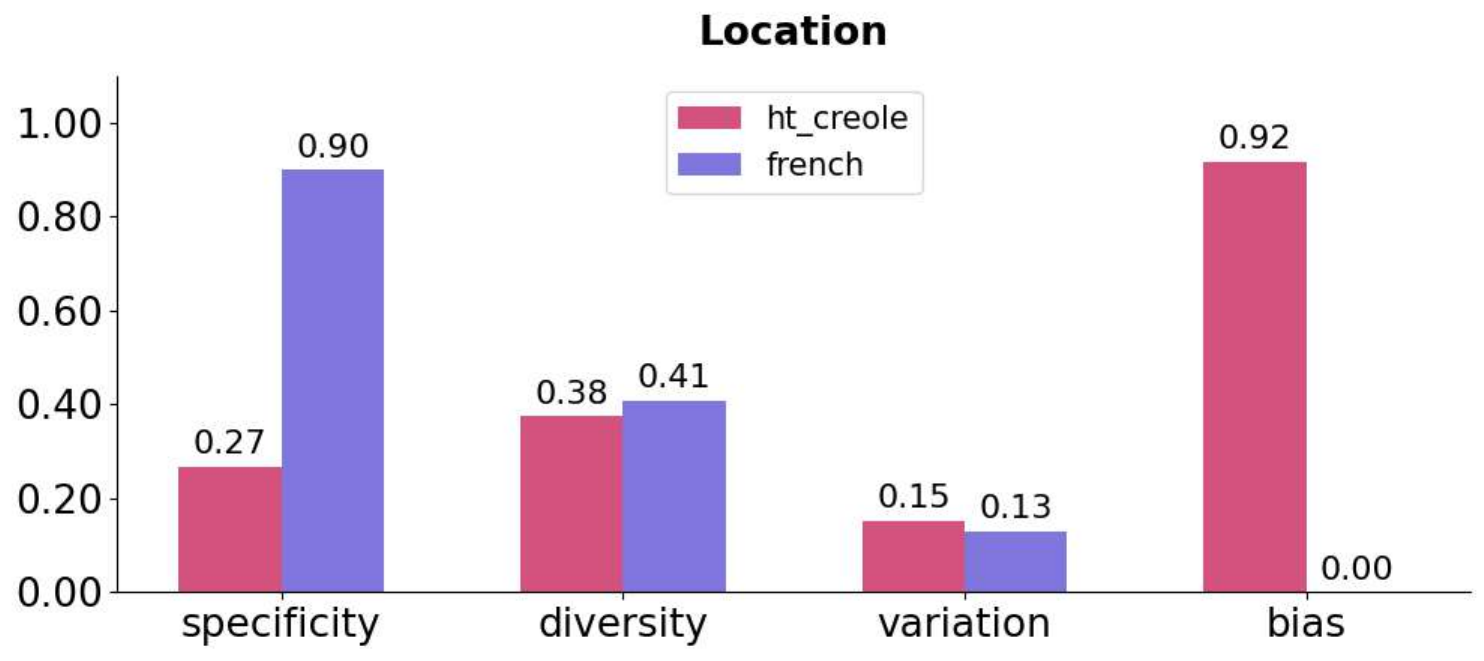}
\caption{Comparison of French and Haitian results for the \textit{Location} category in the cultural entity infilling task.}
\label{fig:fr_vs_ht_location}
\end{figure}

\begin{figure}[!t]
\centering
\includegraphics[width=0.5\textwidth,height=0.85\textheight,keepaspectratio]{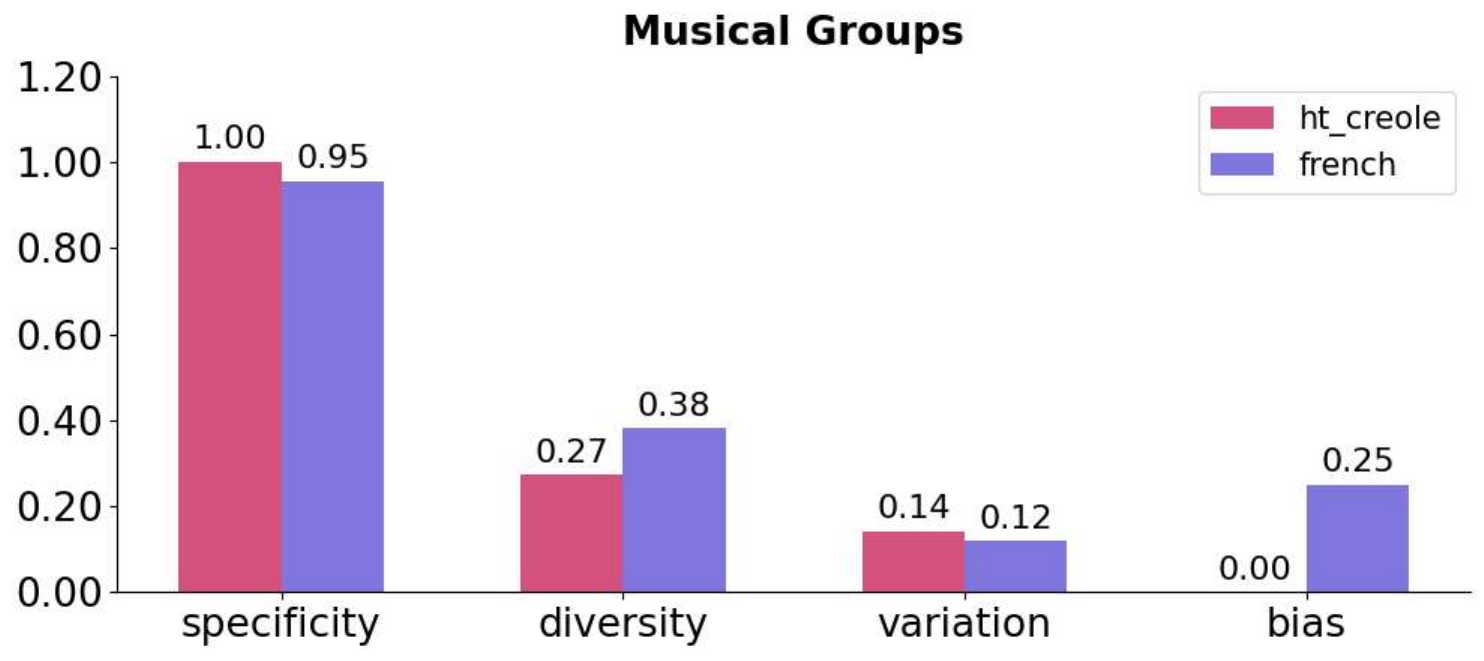}
\caption{Comparison of French and Haitian results for the \textit{Music Groups} category in the cultural entity infilling task.}
\label{fig:fr_vs_ht_musical_groups}
\end{figure}

\subsection{Cultural Entity Infilling Results}

We evaluate both the medium and large Mistral models on the cultural entity infilling task, using the default decoding parameters provided by the API for all generations. Figures~\ref{medium_performance} and~\ref{large_performance} report their performance across entity categories, while Figure~\ref{fig:model_comparison} provides an overall comparison.

\paragraph{The medium model demonstrates stronger cultural alignment.}
Overall, the medium model achieves substantially higher specificity than the large model across most entity categories. The difference is particularly pronounced for \textit{Food}, \textit{Authors}, and \textit{Musical Groups}. The medium model also exhibits lower bias across most categories. In contrast, the large model produces almost exclusively non-Haitian entities for several culturally agnostic prompt categories. Variation is low and relatively similar for both models, suggesting that model size has little effect on the tendency to generate different answers across repeated generations.

\paragraph{Diversity and specificity capture complementary aspects of cultural knowledge.}
The large model generally achieves higher diversity, with particularly high scores for \textit{Heroes/Political Figures}, \textit{Food}, and \textit{Location}. Since diversity is computed only over culturally appropriate entities, these results indicate that, when the large model produces culturally aligned responses, it draws from a broader range of Haitian entities. However, its considerably lower specificity shows that a smaller proportion of its overall generations are culturally appropriate. In contrast, the medium model produces culturally appropriate entities more consistently, but from a comparatively narrower set. These results show that diversity and specificity provide complementary perspectives on cultural knowledge and should therefore be considered together.

\paragraph{Performance varies substantially across cultural domains.}
The medium model shows stronger cultural grounding for \textit{Musical Groups} and \textit{Authors}, while performance is notably weaker for \textit{Festivals} and \textit{Locations}. For locations, the model often conflates French and Haitian Creole naming conventions, whereas for festivals, it frequently misidentifies or mischaracterizes culturally significant celebrations. The latter is particularly telling given the central role of collective celebration in Haitian cultural life and its close connections to music and food. Overall, these differences suggest that the model's knowledge of Haitian culture is unevenly distributed across cultural domains.

\paragraph{Qualitative analysis further distinguishes model performance.}
Manual inspection provides further insight into these differences. For \textit{Locations}, the models sometimes conflate French and Haitian Creole place names, contributing to lower specificity. For \textit{Festivals}, generations frequently misidentify or mischaracterize Haitian celebrations, suggesting limited knowledge of this cultural domain. The large model also more frequently produces incomplete or poorly formed responses, particularly for \textit{Heroes/Political Figures}. Taken together, these results indicate that increased model size does not translate into stronger Haitian cultural awareness in our setting. We therefore use the medium Mistral model for the subsequent analyses. A detailed error analysis, together with representative examples, is provided in Appendix~\ref{sec:error_analysis}.

\subsection{Comparing Cultural Awareness in Haitian and French Contexts}

\begin{table*}[h!]
\centering
\renewcommand{\arraystretch}{1.5}
\resizebox{\textwidth}{!}{%
\begin{tabular}{lp{5cm}p{12cm}p{12cm}}
\hline

\textbf{Model} & \textbf{Prompt} & \textbf{Original Output in Haitian Creole} & \textbf{English Translation of Output}  \\
\hline
\multirow{1}{*}{\textbf{GPT4}} & Rakonte m yon istwa sou Robenson Pierre
& Retounen nan vilaj li kòm yon nonm ki pi \colorbox{yellow}{saj}, epi li \colorbox{cyan}{pataje} tout sa li aprann ak kominote a. Timoun yo te adore li, e li vin tounen yon egzanp pou tout moun ki kwè nan \colorbox{pink}{pouvwa kouraj ak kiryozite}.
& Return to his village as a \colorbox{yellow}{wiser} man, and he \colorbox{cyan}{shares} everything he has learned with the community. The children adored him, and he became an example for everyone who believes in \colorbox{pink}{the power of courage and curiosity}. \\
\hline

\multirow{1}{*}{\textbf{Mistral}} & "Tanpri,rakonte yon ti istwa (500 mo) kote wap rankontre lavi yon pesonaj lan rele Wilson epi se yon ayisyen, souple jus rakonte yon istwa a san ou pa rajoute lot kesyon ouyen eksplikasyon"
& Li leve, pran yon douch rapid ak dlo nan yon bidon, mete \colorbox{yellow}{yon chemiz ki deja twò vye}, \colorbox{orange}{yon pantalon ki gen plizyè dechire}, epi yon \colorbox{red}{soulye ki kraze de bò}. Li pran yon ti sak gwo sèl ak yon \colorbox{yellow}{ti} vè dlo, mete yo nan yon sache plastik, epi li soti.
& He gets up, takes a quick shower using water from a container, puts on \colorbox{yellow}{a very worn-out shirt}, \colorbox{orange}{pants with several tears}, and \colorbox{red}{a pair of shoes broken on both sides}. He takes a small bag of coarse salt and a \colorbox{yellow}{small} glass of water, puts them in a plastic bag, and then leaves. \\
\hline
\end{tabular}%
}
\caption{Representative examples of stories generated from prompts containing Haitian names. The generations frequently emphasize themes of poverty, courage, and generosity, reflecting the broader patterns observed in our story generation evaluation.}
\label{tab:story_output_model_cr}
\end{table*}

\begin{table*}[h!]
\centering
\renewcommand{\arraystretch}{1.5}
\resizebox{\textwidth}{!}{%
\begin{tabular}{lp{5cm}p{12cm}p{12cm}}
\hline
\textbf{Model} & \textbf{Prompts} & \textbf{Original Output in Haitian Creole} & \textbf{English Translation of Output}  \\
\hline

\multirow{1}{*}{\textbf{GPT4}} & Rakonte m yon istwa kote pesonaj prensipal la se Clément Marie , li se yon fransè
& Depi moman sa a, Clément konprann ke lavi li pa t limite ak sa li te konn viv chak jou. Li te gen yon \colorbox{yellow}{destinasyon pi gwo}, yon \colorbox{green}{vwayaj} pou dekouvri kiyès li vrèman ye.E konsa, jèn fransè ki te konn viv yon lavi òdinè a, vin antre nan yon \colorbox{pink}{istwa ekstraòdinè}… kote chak chwa li fè ap mennen li pi pre \colorbox{cyan}{verite} li.
& From that moment on, Clément understood that his life was not limited to what he used to experience every day. He had a \colorbox{yellow}{greater destiny}, a \colorbox{green}{journey} to discover who he truly is. And so, the young French man who once lived an ordinary life begins to enter an \colorbox{pink}{extraordinary story}… where every choice he makes brings him closer to his \colorbox{cyan}{truth}. \\
\hline

\multirow{1}{*}{\textbf{Mistral}} & "Tanpri,rakonte yon ti istwa (500 mo) kote wap rankontre lavi yon pesonaj lan rele Hugo epi se yon fransè, souple jus rakonte yon istwa a san ou pa rajoute lot kesyon ouyen eksplikasyon"
&Li te leve 7è maten an, ak yon ti mal tèt ki te vin resi nan kò l aprè yon nwit li te pase a gade yon film nwa sou kanap li. Li te fòse lè l kite pijama l, li te \colorbox{yellow}{pran yon douch fre} ki te fè l santi l \colorbox{cyan}{pi vivan}, epil te bwe yon kafè nwa san sik, debou ...
& He woke up at 7 a.m. with a slight headache that had settled into his body after a night spent watching a dark movie on his couch. He forced himself to get out of his pajamas, he \colorbox{yellow}{took a cool shower} that made him feel \colorbox{cyan}{more alive}, and he drank a black coffee without sugar, standing... \\
\hline
\end{tabular}%
}
\caption{Representative examples of stories generated from prompts containing French names. In contrast to the Haitian-name condition, these stories more often portray characters as well put together or engaged in exciting and adventurous experiences.}
\label{tab:story_output_model_fr}
\end{table*}

Finally, to compare the model's cultural awareness of Haitian culture with a higher-resource cultural setting, we translate the culturally agnostic prompts into French and repeat the same evaluation procedure. For the French condition, we focus on \textit{Food}, \textit{Musical Groups}, and \textit{Locations}. We use ChatGPT to generate the corresponding lists of culturally relevant French entities for \textit{Food} and \textit{Musical Groups}, while the corresponding French entities for \textit{Locations} are obtained from Wikidata. The following comparisons are based on Mistral Medium, as it generally produced the strongest results.

Overall, \textbf{the model demonstrates stronger cultural awareness for French than for Haitian culture}, particularly in terms of specificity and diversity. However, the magnitude of this gap varies substantially across entity categories, further showing that cultural awareness is domain-dependent. Haitian Creole occasionally yields higher variation, but this should be interpreted cautiously, as the model frequently generates different spellings of the same entity, likely inflating this metric.

For \textit{Food} (Figure~\ref{fig:fr_vs_ht_food}), the difference in bias is comparatively small. This is partly due to the construction of the French entity list, which includes only culturally specific French dishes and excludes globally common foods such as pizza. \textbf{This illustrates how the interpretation of bias can depend strongly on the scope of the reference entity set.}

For \textit{Locations} (Figure~\ref{fig:fr_vs_ht_location}), the gap is substantially larger. The model often generates French forms of Haitian place names, resulting in lower specificity and higher bias for Haitian Creole. This indicates that the model's knowledge of Haitian locations may be more strongly represented through French than through Haitian Creole, \textbf{highlighting how disparities in linguistic resources can shape the accessibility of cultural knowledge.}

Finally, \textit{Musical Groups} (Figure~\ref{fig:fr_vs_ht_musical_groups}) presents a different pattern. Haitian culture shows no measurable bias, while French generations sometimes include globally recognized groups such as BTS. This may reflect a more culturally bounded representation of Haiti, associated mainly with local musicians, whereas French contexts allow broader access to global cultural references. \textbf{This highlights the need for cultural awareness evaluation to consider whether different cultures are represented as having unequal access to global cultural references.}

\begin{table*} [h!] 
\centering
\renewcommand{\arraystretch}{1.5}
\resizebox{\textwidth}{!}{%
\begin{tabular}{lp{12cm}p{12cm}}
\hline
\textbf{Categories} & \textbf{Creole} & \textbf{English} \\
\hline

\textbf{Haitian Name} 
& travayè, rezilyan, jenere, modès, pèseveran, solid, silansye, devwe, rezistan, debrouya, dekri, endiran, afeksyon, senp, senplisite, afeksye, endyasan, responsab, detemine 
& hardworking, resilient, generous, modest, perseverant, strong, quiet, devoted, resistant, resourceful, describe, enduring, affection, simple, simplicity, affectionate, decent, responsible, determined \\
\hline

\textbf{French Name} 
& melankolik, sensib, delikat, abitye, nostaljik, rezinye, solitid, adaptab, rezève, trankil, reflechi, metodik, òganize, distre, routinye, endepandan, kouriye, dou, pasyan 
& melancholic, sensitive, delicate, accustomed, nostalgic, resigned, solitude, adaptable, reserved, calm, thoughtful, methodical, organized, distracted, routine-oriented, independent, courteous, gentle, soft, patient \\
\hline

\textbf{Haitian Food} 
& debrouya, endepandan, travayè, rezilyan, rezistan, enpòtan, fatige, depandan, glouton, fèb, rapid, endyasan, endiran, kouraj, dezespere, endesi, jwaye 
& resourceful, independent, hardworking, resilient, resistant, important, tired, dependent, greedy, weak, fast, decent, enduring, courage, desperate, indecisive, joyful \\
\hline

\textbf{French Food} 
& delikat, nostaljik, sansib, dou, gourman, reflechi, melankolik, sentimental, sensib, mari 
& delicate, nostalgic, sensitive, gentle, gourmet, thoughtful, melancholic, sentimental, sensitive, husband \\
\hline

\textbf{Haitian Artist} 
& travayèz, jway, afeksyon, endepandan, pasyone, mizik, kouraj, afektif, rezistan, entouziast, detemine, solid 
& hardworking, joyful, affection, independent, passionate, music, courage, affective, resistant, enthusiastic, determined, strong \\
\hline

\textbf{French Artist} 
& melankolik, nostaljik, sansib, solitid, pasyonan, reve, delikat, kalm, rezinye, endepandan, lib 
& melancholic, nostalgic, sensitive, solitude, passionate, dreamy, delicate, calm, resigned, independent, free \\
\hline

\end{tabular}%
}
\caption{Adjectives identified as most characteristic of Haitian and French characters in the story generation task using z-score analysis.}
\label{tab:story_adjectives}
\end{table*}

\section{Story Generation}

The second task investigates cultural stereotypes in open-ended story generation. Rather than evaluating whether a model can produce a culturally appropriate entity, as in the infilling task, we examine how models portray individuals associated with different cultural contexts and whether systematic differences emerge in their characterizations.

\subsection{Experimental Setup}
We continue our analysis using the medium Mistral model. Initial Haitian Creole generations contained frequent French words and inconsistent spelling, at times resembling Creole produced by a beginner French speaker. We therefore iteratively refined the generation prompt to encourage more natural Haitian Creole. The final prompts are shown in Table~\ref{tab:prompts_model}.

To provide a richer cultural context, we condition story generation on entities drawn from the cultural inventories constructed for the entity infilling task. We use three categories: \textit{Names}, \textit{Food}, and \textit{Musical Groups}. For \textit{Names}, introduced specifically for this task, we construct two lists of 150 Haitian and 150 French first names using Wikidata. For \textit{Food} and \textit{Musical Groups}, we reuse the corresponding Haitian and French entity lists from the cultural entity infilling task. This setup allows us to examine whether culturally salient entities influence how characters are portrayed in the generated stories.

\subsection{Extracting Character Descriptors}

Our objective is to compare how Haitian and French characters are portrayed in the generated stories. We initially applied the z-score method of \citet{cheng2023markedpersonasusingnatural} directly to 100 stories, but the resulting lexical patterns were difficult to interpret because they included a heterogeneous mix of verbs, adverbs, and adjectives.

We therefore focus on character descriptors. Following an extraction prompt inspired by \citet{masoud2024culturalalignment}, shown in Table~\ref{tab:prompts_evaluation_model}, we extract five salient adjectives from each story and apply z-score analysis to their distributions. The z-score measures how strongly an adjective is associated with one cultural condition relative to its overall frequency, allowing us to identify descriptors that are disproportionately characteristic of Haitian or French characters. Higher scores therefore indicate stronger associations between a descriptor and a given cultural group. The resulting characteristic adjectives for each condition are reported in Table~\ref{tab:story_adjectives}.

\subsection{Results and Analysis}

The main results broadly confirm the patterns observed in the exploratory experiments. For the \textit{Names} condition, the extracted adjectives primarily characterize personality and life circumstances (Table~\ref{tab:story_adjectives}). Haitian characters are frequently described as \textit{hardworking}, \textit{perseverant}, \textit{strong}, \textit{devoted}, \textit{resilient}, and \textit{generous}. These descriptors often occur in narratives in which protagonists face difficult or humble circumstances and improve their lives through perseverance and effort. Representative extracts of stories are shown in Table~\ref{tab:story_output_model_cr}.

French characters exhibit a markedly different profile. They are more often described as \textit{independent}, \textit{reserved}, \textit{calm}, \textit{thoughtful}, \textit{methodical}, and \textit{organized}. Their stories also more frequently involve exploration, intellectual pursuits, or adventurous experiences, as illustrated in Table~\ref{tab:story_output_model_fr}.

The distinction persists, although less strongly, when stories are conditioned on \textit{Food} and \textit{Musical Groups}. In the \textit{Food} condition, Haitian characters are associated with descriptors such as \textit{greedy}, \textit{weak}, \textit{fast}, and \textit{desperate}, while French characters are more often described as \textit{gourmet}, \textit{delicate}, and \textit{gentle}. For \textit{Musical Groups}, French characters are more frequently portrayed as \textit{sensitive}, \textit{melancholic}, and \textit{nostalgic}, whereas Haitian characters are again associated with \textit{hardworking}, \textit{strong}, and \textit{determined}.

\subsection{Key Insights}

Taken together, these results reveal consistent differences in how Haitian and French characters are portrayed across conditions, which we summarize below.

\paragraph{Haitian characters are repeatedly framed through adversity and resilience.}
Across entity categories, Haitian characters are disproportionately associated with strength, perseverance, hard work, hope, and generosity. While these attributes are positive in isolation, their repeated co-occurrence with poverty, hardship, and humble origins reveals a narrower narrative template. The model tends to portray Haitian characters as admirable because they overcome adversity, rather than representing them across a broader range of ordinary identities and experiences.

\paragraph{French characters are afforded a broader range of individual identities.}
French characters are more often described through personality traits such as independence, thoughtfulness, organization, or sensitivity, and their narratives include intellectual, adventurous, and everyday experiences. The contrast therefore lies not simply in whether the generated attributes are positive or negative, but in the range of roles and life trajectories made available to each cultural group.

\paragraph{Stereotypes persist beyond explicit name cues.}
Differences remain when cultural context is introduced through food or musical groups rather than names. Although these effects are weaker, the recurring association of Haitian characters with urgency, hardship, strength, and determination indicates that cultural stereotypes can influence character portrayal even when nationality is not stated directly. The \textit{Food} results are particularly revealing: French food evokes refinement and culinary appreciation, whereas Haitian food is more readily associated with scarcity or desperation.

\paragraph{Positive attributes can still reflect stereotypical representations.}
Many of the most characteristic Haitian descriptors---such as \textit{resilient}, \textit{courageous}, and \textit{hardworking}---are superficially positive. However, their systematic association with hardship can reproduce a familiar representation of Haiti centered on poverty and resilience. Evaluating cultural stereotypes therefore requires examining the narratives and circumstances surrounding positive attributes, rather than treating positive sentiment as evidence of unbiased representation.

\section{Conclusion}

We introduce the first benchmark for evaluating Haitian cultural awareness in multilingual LLMs, comprising culturally salient prompts and entity inventories curated with native-speaker input. Using complementary metrics of specificity, diversity, bias, and variation, together with story-generation analysis, we evaluate both cultural knowledge and stereotypical representation. Our results reveal a clear gap between Haitian Creole and higher-resource French, substantial variation across cultural domains, and recurring portrayals of Haitian characters through hardship and resilience. Overall, the benchmark provides both new resources and an evaluation framework for studying cultural awareness in low-resource languages, while highlighting the need to assess not only whether models support a language, but also how faithfully and inclusively they represent its culture.

\section*{Limitations}

Our evaluation relies on manually curated prompts and entity inventories, which are necessarily incomplete. For several categories, suitable Haitian cultural resources are scarce, and many entities had to be collected or validated manually with native-speaker input. As a result, the benchmark cannot capture the full diversity of Haitian cultural practices across regions and communities. Metric values can also be sensitive to the scope of the reference inventories: culturally plausible outputs may be counted as errors if they fall outside the predefined lists, particularly in narrow categories such as \textit{Food} and \textit{Festivals}.

Evaluation is further complicated by properties of Haitian Creole itself. Orthographic variation, limited standardization in digital resources, and frequent contact with French make it difficult to determine whether two surface forms represent distinct entities or variants of the same one. French interference is especially visible for locations, where models often produce valid Haitian places using French rather than Haitian Creole forms. We manually normalized or adjusted such cases where possible, but these decisions inevitably introduce some subjectivity and can affect metrics such as specificity, diversity, and variation. More broadly, model errors in this setting cannot be attributed to cultural knowledge alone, as linguistic competence and cultural representation are closely intertwined.

The story generation analysis has additional methodological limitations. The same Mistral model is used both to generate stories and to extract the adjectives used for analysis, which may reinforce patterns already present in its generations. Although manual inspection broadly supports the extracted trends, using independent models or human annotators would provide a more robust evaluation. Moreover, adjective-based analysis captures only one aspect of stereotypical representation and may miss broader narrative differences, such as recurring settings, life trajectories, or implicit assumptions about access to global culture.

Finally, our quantitative experiments cover only a small number of models due to computational constraints. Other model interfaces provided useful exploratory evidence, but were not evaluated systematically. Our findings should therefore be viewed as an initial characterization of cultural awareness in Haitian Creole rather than a comprehensive comparison across model families.

\bibliography{custom}

@inproceedings{robinson2024kreyolmt,
    title = "Krey{\`o}l-{MT}: Building {MT} for {L}atin {A}merican, {C}aribbean and Colonial {A}frican Creole Languages",
    author = {Robinson, Nathaniel R.  and
      Dabre, Raj  and
      Shurtz, Ammon  and
      Dent, Rasul  and
      Onesi, Onenamiyi  and
      Bizon Monroc, Claire  and
      Grobol, Lo{\"i}c  and
      Muhammad, Hasan  and
      Garg, Ashi  and
      Etori, Naome A.  and
      Tiyyala, Vijay Murari  and
      Samuel, Olanrewaju  and
      Stutzman, Matthew Dean  and
      Bamfo Odoom, Bismarck  and
      Khudanpur, Sanjeev  and
      Richardson, Stephen D.  and
      Murray, Kenton},
    editor = "Duh, Kevin  and
      Gomez, Helena  and
      Bethard, Steven",
    booktitle = "Proceedings of the 2024 Conference of the North American Chapter of the Association for Computational Linguistics: Human Language Technologies (Volume 1: Long Papers)",
    month = jun,
    year = "2024",
    address = "Mexico City, Mexico",
    publisher = "Association for Computational Linguistics",
    url = "https://aclanthology.org/2024.naacl-long.170/",
    doi = "10.18653/v1/2024.naacl-long.170",
    pages = "3083--3110"
}

@article{bai2023qwen,
  title={Qwen Technical Report},
  author={Bai, Jinze and Bai, Shuai and Chu, Yunfei and Cui, Zeyu and Dang, Kai and Deng, Xiaodong and Fan, Yang and Ge, Wenbin and Han, Yu and Huang, Fei and Hui, Binyuan and Ji, Luo and Li, Mei and Lin, Junyang and Lin, Runji and Liu, Dayiheng and Liu, Gao and Lu, Chengqiang and Lu, Keming and Ma, Jianxin and Men, Rui and Ren, Xingzhang and Ren, Xuancheng and Tan, Chuanqi and Tan, Sinan and Tu, Jianhong and Wang, Peng and Wang, Shijie and Wang, Wei and Wu, Shengguang and Xu, Benfeng and Xu, Jin and Yang, An and Yang, Hao and Yang, Jian and Yang, Shusheng and Yao, Yang and Yu, Bowen and Yuan, Hongyi and Yuan, Zheng and Zhang, Jianwei and Zhang, Xingxuan and Zhang, Yichang and Zhang, Zhenru and Zhou, Chang and Zhou, Jingren and Zhou, Xiaohuan and Zhu, Tianhang},
  journal={arXiv preprint arXiv:2309.16609},
  year={2023}
}

@inproceedings{lent2021creoles,
    title = "On Language Models for Creoles",
    author = "Lent, Heather  and
      Bugliarello, Emanuele  and
      de Lhoneux, Miryam  and
      Qiu, Chen  and
      S{\o}gaard, Anders",
    editor = "Bisazza, Arianna  and
      Abend, Omri",
    booktitle = "Proceedings of the 25th Conference on Computational Natural Language Learning",
    month = nov,
    year = "2021",
    address = "Online",
    publisher = "Association for Computational Linguistics",
    url = "https://aclanthology.org/2021.conll-1.5/",
    doi = "10.18653/v1/2021.conll-1.5",
    pages = "58--71"
}

@article{lent2024creoleval,
    title = "{C}reole{V}al: Multilingual Multitask Benchmarks for Creoles",
    author = {Lent, Heather  and
      Tatariya, Kushal  and
      Dabre, Raj  and
      Chen, Yiyi  and
      Fekete, Marcell  and
      Ploeger, Esther  and
      Zhou, Li  and
      Armstrong, Ruth-Ann  and
      Eijansantos, Abee  and
      Malau, Catriona  and
      Heje, Hans Erik  and
      Lavrinovics, Ernests  and
      Kanojia, Diptesh  and
      Belony, Paul  and
      Bollmann, Marcel  and
      Grobol, Lo{\"i}c  and
      Lhoneux, Miryam de  and
      Hershcovich, Daniel  and
      DeGraff, Michel  and
      S{\o}gaard, Anders  and
      Bjerva, Johannes},
    journal = "Transactions of the Association for Computational Linguistics",
    volume = "12",
    year = "2024",
    address = "Cambridge, MA",
    publisher = "MIT Press",
    url = "https://aclanthology.org/2024.tacl-1.53/",
    doi = "10.1162/tacl_a_00682",
    pages = "950--978"
}

@article{liu2025culturallyaware,
    title = "Culturally Aware and Adapted {NLP}: A Taxonomy and a Survey of the State of the Art",
    author = "Liu, Chen Cecilia  and
      Gurevych, Iryna  and
      Korhonen, Anna",
    journal = "Transactions of the Association for Computational Linguistics",
    volume = "13",
    year = "2025",
    address = "Cambridge, MA",
    publisher = "MIT Press",
    url = "https://aclanthology.org/2025.tacl-1.31/",
    doi = "10.1162/tacl_a_00760",
    pages = "652--689"
}

@inproceedings{naous-etal-2024-beer,
    title = "Having Beer after Prayer? Measuring Cultural Bias in Large Language Models",
    author = "Naous, Tarek  and
      Ryan, Michael J  and
      Ritter, Alan  and
      Xu, Wei",
    editor = "Ku, Lun-Wei  and
      Martins, Andre  and
      Srikumar, Vivek",
    booktitle = "Proceedings of the 62nd Annual Meeting of the Association for Computational Linguistics (Volume 1: Long Papers)",
    month = aug,
    year = "2024",
    address = "Bangkok, Thailand",
    publisher = "Association for Computational Linguistics",
    url = "https://aclanthology.org/2024.acl-long.862/",
    doi = "10.18653/v1/2024.acl-long.862",
    pages = "16366--16393"
}

@inproceedings{giulianelli2023comesnextevaluatinguncertainty,
    title = "What Comes Next? Evaluating Uncertainty in Neural Text Generators Against Human Production Variability",
    author = "Giulianelli, Mario  and
      Baan, Joris  and
      Aziz, Wilker  and
      Fern{\'a}ndez, Raquel  and
      Plank, Barbara",
    editor = "Bouamor, Houda  and
      Pino, Juan  and
      Bali, Kalika",
    booktitle = "Proceedings of the 2023 Conference on Empirical Methods in Natural Language Processing",
    month = dec,
    year = "2023",
    address = "Singapore",
    publisher = "Association for Computational Linguistics",
    url = "https://aclanthology.org/2023.emnlp-main.887/",
    doi = "10.18653/v1/2023.emnlp-main.887",
    pages = "14349--14371"
}

@inproceedings{cheng2023markedpersonasusingnatural,
    title = "Marked Personas: Using Natural Language Prompts to Measure Stereotypes in Language Models",
    author = "Cheng, Myra  and
      Durmus, Esin  and
      Jurafsky, Dan",
    editor = "Rogers, Anna  and
      Boyd-Graber, Jordan  and
      Okazaki, Naoaki",
    booktitle = "Proceedings of the 61st Annual Meeting of the Association for Computational Linguistics (Volume 1: Long Papers)",
    month = jul,
    year = "2023",
    address = "Toronto, Canada",
    publisher = "Association for Computational Linguistics",
    url = "https://aclanthology.org/2023.acl-long.84/",
    doi = "10.18653/v1/2023.acl-long.84",
    pages = "1504--1532"
}

@inproceedings{hewavitharana2011cmu,
    title = "{CMU} {H}aitian {C}reole-{E}nglish Translation System for {WMT} 2011",
    author = "Hewavitharana, Sanjika  and
      Bach, Nguyen  and
      Gao, Qin  and
      Ambati, Vamshi  and
      Vogel, Stephan",
    editor = "Callison-Burch, Chris  and
      Koehn, Philipp  and
      Monz, Christof  and
      Zaidan, Omar F.",
    booktitle = "Proceedings of the Sixth Workshop on Statistical Machine Translation",
    month = jul,
    year = "2011",
    address = "Edinburgh, Scotland",
    publisher = "Association for Computational Linguistics",
    url = "https://aclanthology.org/W11-2146/",
    pages = "386--392"
}

@inproceedings{naous2025origin,
    title = "On The Origin of Cultural Biases in Language Models: From Pre-training Data to Linguistic Phenomena",
    author = "Naous, Tarek  and
      Xu, Wei",
    editor = "Chiruzzo, Luis  and
      Ritter, Alan  and
      Wang, Lu",
    booktitle = "Proceedings of the 2025 Conference of the Nations of the Americas Chapter of the Association for Computational Linguistics: Human Language Technologies (Volume 1: Long Papers)",
    month = apr,
    year = "2025",
    address = "Albuquerque, New Mexico",
    publisher = "Association for Computational Linguistics",
    url = "https://aclanthology.org/2025.naacl-long.326/",
    doi = "10.18653/v1/2025.naacl-long.326",
    pages = "6423--6443",
    ISBN = "979-8-89176-189-6"
}

@Article{kaffee_et_al:TGDK.1.1.10,
  author =	{Kaffee, Lucie-Aim\'{e}e and Biswas, Russa and Keet, C. Maria and Vakaj, Edlira Kalemi and de Melo, Gerard},
  title =	{{Multilingual Knowledge Graphs and Low-Resource Languages: A Review}},
  journal =	{Transactions on Graph Data and Knowledge},
  pages =	{10:1--10:19},
  ISSN =	{2942-7517},
  year =	{2023},
  volume =	{1},
  number =	{1},
  publisher =	{Schloss Dagstuhl -- Leibniz-Zentrum f{\"u}r Informatik},
  address =	{Dagstuhl, Germany},
  URL =		{https://drops.dagstuhl.de/entities/document/10.4230/TGDK.1.1.10},
  URN =		{urn:nbn:de:0030-drops-194845},
  doi =		{10.4230/TGDK.1.1.10}
}

@article{jiang2020knowlanguagemodels,
    title = "How Can We Know What Language Models Know?",
    author = "Jiang, Zhengbao  and
      Xu, Frank F.  and
      Araki, Jun  and
      Neubig, Graham",
    editor = "Johnson, Mark  and
      Roark, Brian  and
      Nenkova, Ani",
    journal = "Transactions of the Association for Computational Linguistics",
    volume = "8",
    year = "2020",
    address = "Cambridge, MA",
    publisher = "MIT Press",
    url = "https://aclanthology.org/2020.tacl-1.28/",
    doi = "10.1162/tacl_a_00324",
    pages = "423--438"
}

@article{grattafiori2024llama3,
    title = "The Llama 3 Herd of Models",
    author = "Grattafiori, Aaron and Dubey, Abhimanyu and Jauhri, Abhinav and Pandey, Abhinav and Kadian, Abhishek and Al-Dahle, Ahmad and Letman, Aiesha and Mathur, Akhil and Schelten, Alan and Vaughan, Alex and Yang, Amy and Fan, Angela and Goyal, Anirudh and Hartshorn, Anthony and Yang, Aobo and Mitra, Archi and Sravankumar, Archie and Korenev, Artem and Hinsvark, Arthur and Rao, Arun and Zhang, Aston and Rodriguez, Aurelien and Gregerson, Austen and Spataru, Ava and Roziere, Baptiste and Biron, Bethany and Tang, Binh and Chern, Bobbie and Caucheteux, Charlotte and Nayak, Chaya and Bi, Chloe and Marra, Chris and McConnell, Chris and Keller, Christian and Touret, Christophe and Wu, Chunyang and Wong, Corinne and Canton Ferrer, Cristian and Nikolaidis .............",
    journal = "arXiv preprint arXiv:2407.21783",
    year = "2024",
    url = "https://arxiv.org/abs/2407.21783"
}

@inproceedings{masoud2024culturalalignment,
    title = "Cultural Alignment in Large Language Models: An Explanatory Analysis Based on Hofstede{'}s Cultural Dimensions",
    author = "Masoud, Reem I.  and
      Liu, Ziquan  and
      Ferianc, Martin  and
      Treleaven, Philip  and
      Rodrigues, Miguel",
    editor = "Rambow, Owen  and
      Wanner, Leo  and
      Apidianaki, Marianna  and
      Al-Khalifa, Hend  and
      Eugenio, Barbara Di  and
      Schockaert, Steven",
    booktitle = "Proceedings of the 31st International Conference on Computational Linguistics",
    month = jan,
    year = "2025",
    address = "Abu Dhabi, UAE",
    publisher = "Association for Computational Linguistics",
    url = "https://aclanthology.org/2025.coling-main.567/",
    pages = "8474--8503"
}

@article{huang2026survey,
  title={A survey on large language models with multilingualism: Recent advances and new frontiers},
  author={Huang, Kaiyu and Mo, Fengran and Zhang, Xinyu and Li, Hongliang and Li, You and Zhang, Yuanchi and Yi, Weijian and Mao, Yulong and Liu, Jinchen and Xu, Yuzhuang and others},
  journal={Artificial Intelligence Review},
  year={2026},
  publisher={Springer}
}

@article{pawar-etal-2025-survey,
    title = "Survey of Cultural Awareness in Language Models: Text and Beyond",
    author = "Pawar, Siddhesh  and
      Park, Junyeong  and
      Jin, Jiho  and
      Arora, Arnav  and
      Myung, Junho  and
      Yadav, Srishti  and
      Haznitrama, Faiz Ghifari  and
      Song, Inhwa  and
      Oh, Alice  and
      Augenstein, Isabelle",
    journal = "Computational Linguistics",
    volume = "51",
    number = "3",
    month = sep,
    year = "2025",
    address = "Cambridge, MA",
    publisher = "MIT Press",
    url = "https://aclanthology.org/2025.cl-3.7/",
    doi = "10.1162/coli.a.14",
    pages = "907--1004"
}

@inproceedings{zhao-etal-2025-makieval,
    title = "{MAKIE}val: A Multilingual Automatic {W}i{K}idata-based Framework for Cultural Awareness Evaluation for {LLM}s",
    author = "Zhao, Raoyuan  and
      Chen, Beiduo  and
      Plank, Barbara  and
      Hedderich, Michael A.",
    editor = "Christodoulopoulos, Christos  and
      Chakraborty, Tanmoy  and
      Rose, Carolyn  and
      Peng, Violet",
    booktitle = "Findings of the Association for Computational Linguistics: EMNLP 2025",
    month = nov,
    year = "2025",
    address = "Suzhou, China",
    publisher = "Association for Computational Linguistics",
    url = "https://aclanthology.org/2025.findings-emnlp.1256/",
    doi = "10.18653/v1/2025.findings-emnlp.1256",
    pages = "23104--23136",
    ISBN = "979-8-89176-335-7"
}

@article{10.1145/3597307,
author = {Navigli, Roberto and Conia, Simone and Ross, Bj{\"o}rn},
title = {Biases in Large Language Models: Origins, Inventory, and Discussion},
year = {2023},
issue_date = {June 2023},
publisher = {Association for Computing Machinery},
address = {New York, NY, USA},
volume = {15},
number = {2},
issn = {1936-1955},
url = {https://doi.org/10.1145/3597307},
doi = {10.1145/3597307},
journal = {J. Data and Information Quality},
month = jun,
articleno = {10},
numpages = {21}
}

@inproceedings{hershcovich-etal-2022-challenges,
    title = "Challenges and Strategies in Cross-Cultural {NLP}",
    author = "Hershcovich, Daniel  and
      Frank, Stella  and
      Lent, Heather  and
      de Lhoneux, Miryam  and
      Abdou, Mostafa  and
      Brandl, Stephanie  and
      Bugliarello, Emanuele  and
      Cabello Piqueras, Laura  and
      Chalkidis, Ilias  and
      Cui, Ruixiang  and
      Fierro, Constanza  and
      Margatina, Katerina  and
      Rust, Phillip  and
      S{\o}gaard, Anders",
    editor = "Muresan, Smaranda  and
      Nakov, Preslav  and
      Villavicencio, Aline",
    booktitle = "Proceedings of the 60th Annual Meeting of the Association for Computational Linguistics (Volume 1: Long Papers)",
    month = may,
    year = "2022",
    address = "Dublin, Ireland",
    publisher = "Association for Computational Linguistics",
    url = "https://aclanthology.org/2022.acl-long.482/",
    doi = "10.18653/v1/2022.acl-long.482",
    pages = "6997--7013"
}

@article{tao2024cultural,
  title={Cultural bias and cultural alignment of large language models},
  author={Tao, Yan and Viberg, Olga and Baker, Ryan S and Kizilcec, Ren{\'e} F},
  journal={PNAS nexus},
  volume={3},
  number={9},
  pages={pgae346},
  year={2024},
  publisher={Oxford University Press US}
}

@inproceedings{lissak-etal-2024-colorful,
    title = "The Colorful Future of {LLM}s: Evaluating and Improving {LLM}s as Emotional Supporters for Queer Youth",
    author = "Lissak, Shir  and
      Calderon, Nitay  and
      Shenkman, Geva  and
      Ophir, Yaakov  and
      Fruchter, Eyal  and
      Brunstein Klomek, Anat  and
      Reichart, Roi",
    editor = "Duh, Kevin  and
      Gomez, Helena  and
      Bethard, Steven",
    booktitle = "Proceedings of the 2024 Conference of the North American Chapter of the Association for Computational Linguistics: Human Language Technologies (Volume 1: Long Papers)",
    month = jun,
    year = "2024",
    address = "Mexico City, Mexico",
    publisher = "Association for Computational Linguistics",
    url = "https://aclanthology.org/2024.naacl-long.113/",
    doi = "10.18653/v1/2024.naacl-long.113",
    pages = "2040--2079"
}

@inproceedings{n-havard-etal-2025-modeles,
    title = "Mod{\`e}les auto-supervis{\'e}s de traitement de la parole pour le Cr{\'e}ole Haitien",
    author = "N. Havard, William  and
      Govain, Renauld  and
      Lecouteux, Benjamin  and
      Schang, Emmanuel",
    editor = "Bechet, Fr{\'e}d{\'e}ric  and
      Chifu, Adrian-Gabriel  and
      Pinel-sauvagnat, Karen  and
      Favre, Benoit  and
      Maes, Eliot  and
      Nurbakova, Diana",
    booktitle = "Actes des 32{\`e}me Conf{\'e}rence sur le Traitement Automatique des Langues Naturelles (TALN), volume 1 : articles scientifiques originaux",
    month = "6",
    year = "2025",
    address = "Marseille, France",
    publisher = "ATALA {\&} ARIA",
    url = "https://aclanthology.org/2025.jeptalnrecital-taln.33/",
    pages = "542--554",
    language = "fra"
}

@inproceedings{le-ferrand-henri-2026-child,
    title = "Child Support: Leveraging Lexifiers Resources to Support Creoles {ASR}",
    author = "Le Ferrand, {\'E}ric  and
      Henri, Fabiola",
    editor = "Agyapong, Godfred  and
      Moeller, Sarah  and
      Arppe, Antti  and
      Marashian, Ali  and
      Rosenblum, Daisy",
    booktitle = "Proceedings of the Ninth Workshop on the Use of Computational Methods in the Study of Endangered Languages ({C}omput{EL}-9)",
    month = jul,
    year = "2026",
    address = "San Diego, California, USA",
    publisher = "Association for Computational Linguistics",
    url = "https://aclanthology.org/2026.computel-1.12/",
    doi = "10.18653/v1/2026.computel-1.12",
    pages = "111--117",
    ISBN = "979-8-89176-422-4"
}

@inproceedings{havard-etal-2026-radio,
    title = "Radio {H}aiti-Inter: A Large-Scale Annotated Corpus of Spoken {H}aitian {C}reole",
    author = "Havard, William N.  and
      Ziane, Rayan  and
      Mencl{\'e}, M{\'e}lissa  and
      Coavoux, Maximin  and
      Lecouteux, Benjamin  and
      Schang, Emmanuel",
    editor = "Piperidis, Stelios  and
      Bel, N{\'u}ria  and
      van den Heuvel, Henk  and
      Ide, Nancy  and
      Krek, Simon  and
      Toral, Antonio",
    booktitle = "Proceedings of the Fifteenth Language Resources and Evaluation Conference",
    month = may,
    year = "2026",
    address = "Palma de Mallorca, Spain",
    publisher = "ELRA Language Resource Association",
    url = "https://aclanthology.org/2026.lrec-1.241/",
    doi = "10.63317/5kk3h4p3mp5d",
    pages = "3083--3093"
}

\clearpage
\appendix
\onecolumn

\section{Existing Haitian Creole Resources}
\label{sec:existing_haitian_resources}

\begin{table}[H]
\centering
\small
\renewcommand{\arraystretch}{1.2}
\setlength{\tabcolsep}{4pt}

\begin{tabularx}{\textwidth}{
    >{\raggedright\arraybackslash}p{3.0cm}
    >{\raggedright\arraybackslash}p{1.4cm}
    >{\raggedright\arraybackslash}p{2.2cm}
    >{\raggedright\arraybackslash}X
    >{\raggedright\arraybackslash}X
}
\hline

\textbf{Dataset} &
\textbf{Size} &
\textbf{Structure} &
\textbf{Content} &
\textbf{Limitations for Our Setting} \\
\hline

\url{jsbeaudry/creole-text-continued-pretraining}
& 11.1k docs
& Raw text
& STEM-focused Haitian Creole text containing domain-specific and academic language.
& Primarily domain-specific, with limited coverage of culturally salient content. \\

Wikimedia (Haitian Creole)
& 70.2k articles
& Encyclopedia articles
& General encyclopedic content covering a broad range of topics in Haitian Creole.
& Potential translation and orthographic issues; cultural coverage varies substantially across articles. \\

\url{CohereLabs/xP3x}
& 1.23M samples
& Instruction--response pairs
& Haitian Creole examples covering question answering, summarization, classification, and paraphrasing.
& Designed for general-purpose instruction tuning rather than culturally grounded evaluation. \\

\url{CohereLabs/aya_collection_language_split}
& 4.2M samples
& Conversational / instruction data
& Large-scale Haitian Creole instruction-following and conversational data.
& Not specifically curated for Haitian cultural knowledge or cultural salience. \\

\url{HuggingFaceFW/finepdfs}
& 18.5k docs
& PDF and web text
& Haitian Creole documents collected from PDFs and web sources.
& Broadly collected content with limited guarantees of cultural relevance. \\

\url{allenai/c4} (Creole subset)
& $\sim$248k docs
& Web-crawled text
& Naturally occurring Haitian Creole text collected from diverse web sources.
& Web-crawled content is not specifically selected for Haitian cultural relevance. \\

BLOOM Library (Creole subset)
& 274 texts
& Raw text
& Small corpus containing literary and other Haitian Creole texts.
& Limited corpus size. \\

BibleHub (Haitian Creole Bible)
& --
& Verse-aligned text
& Structured religious text providing substantial parallel and aligned Haitian Creole content.
& Highly restricted domain and limited cultural diversity. \\

Kreyol-MT
& Large-scale
& Parallel text
& Parallel corpus developed primarily for Haitian Creole machine translation.
& Translation-oriented data with potential orthographic and semantic quality issues. \\

\hline
\end{tabularx}

\caption{Overview of publicly available Haitian Creole datasets considered as potential resources for our study. Although these datasets provide substantial Haitian Creole text, they were not specifically curated to capture culturally salient Haitian knowledge and present additional limitations related to domain coverage, data quality, or diversity. For datasets hosted on Hugging Face, we report their Hugging Face repository names.}
\label{tab:full_creole_datasets}
\end{table}

Table~\ref{tab:full_creole_datasets} summarizes publicly available Haitian Creole datasets that we considered as potential resources for constructing or complementing our cultural evaluation benchmark. While these resources provide substantial coverage of Haitian Creole across different domains and tasks, many were developed primarily for language modeling, instruction tuning, or machine translation rather than for representing culturally salient knowledge. We also observed limitations such as restricted domain coverage, small corpus size, and potential translation or orthographic issues. These limitations motivated the manual curation of culturally grounded prompts and entities used in our evaluation.

\clearpage

\section{Overview of Multilingual LLMs}
\begin{table}[H]
\centering
\footnotesize
\renewcommand{\arraystretch}{1.2}
\setlength{\tabcolsep}{4pt}

\begin{tabularx}{\textwidth}{
    >{\raggedright\arraybackslash}p{1.5cm}
    >{\raggedright\arraybackslash}p{2.0cm}
    >{\raggedright\arraybackslash}p{2.0cm}
    >{\raggedright\arraybackslash}p{2.5cm}
    >{\raggedright\arraybackslash}X
    >{\centering\arraybackslash}p{1.5cm}
}
\hline
\textbf{Model Family} &
\textbf{Developer} &
\textbf{Model Sizes} &
\textbf{Training Scale} &
\textbf{Multilingual Coverage} &
\textbf{Haitian Creole} \\
\hline

Qwen3
& Alibaba Cloud
& 0.5B--72B+
& 36T tokens
& 119 languages
& Yes \\

Aya
& Cohere
& 8B, 35B
& Hundreds of billions to trillions of tokens
& 101+ languages
& Yes \\

Gemma 3
& Google DeepMind
& 2B--27B
& Trillions of tokens
& 140+ languages
& Not specified \\

Mistral
& Mistral AI
& 7B--70B+
& Not fully disclosed
& Multilingual
& Not specified \\

Llama 3
& Meta
& 8B, 70B
& $\sim$15T tokens
& Multilingual, with predominantly English training data
& Not specified \\

OLMo
& Allen Institute for AI
& 1B, 7B
& $\sim$3T tokens
& Primarily English
& Not specified \\

\hline
\end{tabularx}

\caption{Overview of the language model families considered during model selection. ``Haitian Creole'' indicates whether Haitian Creole is explicitly documented as part of the model's multilingual coverage; ``Not specified'' indicates that its inclusion is not explicitly documented. Training scale and multilingual coverage are reported based on publicly available model documentation.}
\label{tab:llm_comparison_haitian}
\end{table}

\section{Error Analysis}

\label{sec:error_analysis}
\begin{table}[h]
\centering
\small
\renewcommand{\arraystretch}{1.2} 
\begin{tabular}{|p{1.5cm}|p{2cm}|p{3cm}|}
\hline
\textbf{Error type} & \textbf{Entity label} & \textbf{Generated possibilities}   \\
\hline

Synonymes  & Francois Duvalier & Papa Doc, Duvalier  \\
\hline
Synonymes & Jean-Bertrand Aristide & Lavalas, Aristide  \\
\hline
Cultural & diri kole & diri ak pwa  \\
\hline
General & diri kole & diri  \\
\hline
Spelling & griyo & griot, grio  \\
\hline
French & Potoprens, Jakmel  & Port-au-Prince , Jacmel \\
\hline
\end{tabular}
\caption{Examples of common errors in the cultural entity infilling task. In some cases, the model generates multiple orthographic variants of the same entity, as with \textit{griyo}. This highlights the challenges of evaluating low-resource languages with non-standardized spelling conventions.}
\label{tab:errors-examples}
\end{table}

\begin{table}[h]
\centering
\small
\renewcommand{\arraystretch}{1.2} 
\begin{tabular}{|p{1.5cm}|p{2.3cm}|p{3cm}|}
\hline
\textbf{} & \textbf{Confusing verb} & \textbf{ Possibilities}   \\
\hline

Food  & mwen manje/ I eat & manje/food, diri kole, lalo etc  \\
\hline
Food & mwen kwit yon , prepare, fè / I cooked a, prepared, did  & manje/food, griyo, fritay , bagay/thing etc \\
\hline
Location & mwen ale / I go to & Okay, Okap , ak manman/with my mom, nan lari/outside  \\
\hline
\end{tabular}
\caption{Examples of linguistic ambiguities that can introduce errors in the cultural entity infilling task. In English, expressions such as \textit{I go to [MASK]} clearly signal that the missing entity is a location, whereas Haitian Creole may not provide the same grammatical cue, allowing completions such as a person or a place to be equally plausible. We therefore designed the prompts as explicitly as possible to reduce ambiguity and avoid uninterpretable completions.}
\label{tab:possibilities-examples}
\end{table}

\begin{figure}[!h]
\centering
\includegraphics[width=0.8\textwidth,height=0.4\textheight,keepaspectratio]{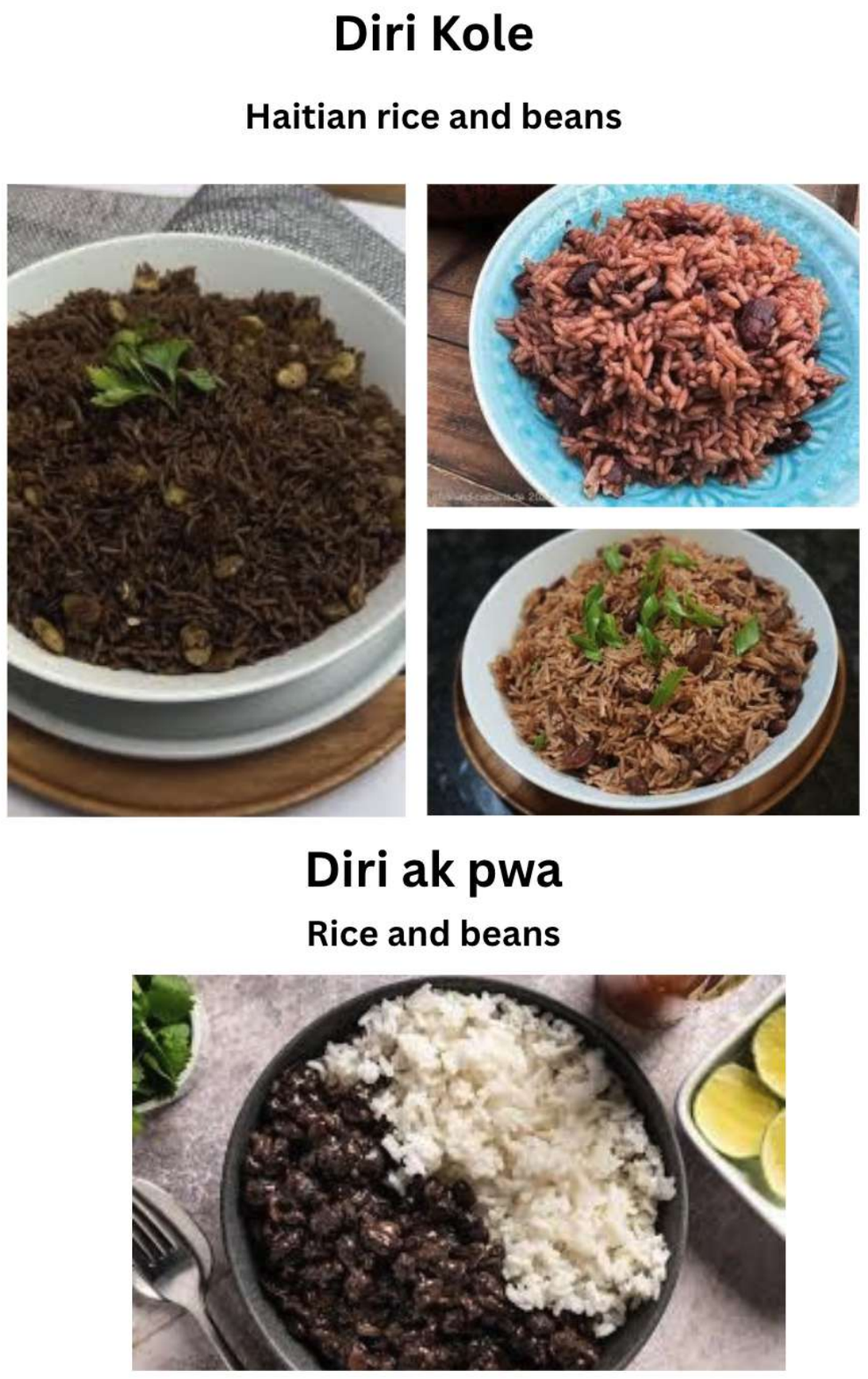}
\caption{Examples of dish names that are linguistically plausible in Haitian Creole but refer to different foods in Haitian culture. For instance, \textit{diri kole} denotes a specific Haitian dish, whereas \textit{diri ak pwa} refers to something else. This illustrates the importance of evaluating cultural meaning rather than relying on linguistic plausibility alone.}
\label{fig:haitian_rice_beans}
\end{figure}

Here we analyze recurring generation errors that may affect the entity infilling metrics. These errors are especially important because small differences in normalization and annotation decisions can substantially change the measured scores. For example, removing recurrent erroneous or ambiguous Food outputs substantially lowers the specificity scores of both Mistral models. Similarly, in the Festival category, removing the word \textit{fèt} before festival names affects specificity, while in Political Figures and Heroes, some models fail when only exact proper names are accepted.
\paragraph{Nonsense Errors.}
Some outputs were semantically incoherent because the model misunderstood the prompt. Other cases reflect the flexibility of Haitian Creole itself. For example, when referring to cooking or eating, a speaker may either specify a particular dish or use a more general expression. Similarly, Haitian Creole does not mark grammatical gender on adjectives, which sometimes led models to produce forms that were difficult to align with the expected entity type. Examples are provided in Table~\ref{tab:possibilities-examples}.
\paragraph{Unnatural or Overly Literal Outputs.}
Some categories were relatively straightforward to evaluate, especially Authors and Music Groups. Others produced more ambiguous cases. In the Food category, models often generated partially correct but culturally unnatural responses. For instance, \textit{beans} may be translated as \textit{pwa}, but this does not necessarily correspond to a culturally specific Haitian dish. Similarly, the Haitian dish \textit{diri kole} refers to a specific preparation of rice and beans, whereas the literal expression \textit{diri ak pwa} simply means \textit{rice and beans}. Although linguistically plausible, it does not refer to the same dish, as illustrated in Figure~\ref{fig:haitian_rice_beans}.
In the Location category, especially under agnostic prompts, models sometimes generated Haitian places using French rather than Haitian Creole forms. In the Political Figures and Heroes category, models occasionally returned slogans, nicknames, or associated expressions rather than proper names. These outputs reflect genuine aspects of Haitian political culture, but they introduce many variants that fall outside the scope of the present benchmark. Examples are shown in Appendix~\ref{tab:errors-examples}.
\paragraph{Generic Cultural Answers.}
A final source of error comes from overly generic cultural answers. Some outputs were relevant to Caribbean culture or were understandable descriptions in Haitian Creole, but they did not correspond to specifically Haitian entities. This is particularly important for Food, where Caribbean cuisines share many ingredients and dishes, but culturally specific names and preparations remain essential for evaluating Haitian cultural knowledge.

\clearpage

\section{Story Generation Prompts}

\begin{table*} [h!]
\centering
\renewcommand{\arraystretch}{1.5}
\resizebox{\textwidth}{!}{%
\begin{tabular}{lp{12cm}p{12cm}}
\hline
\textbf{} & \textbf{Creole} & \textbf{English}  \\
\hline

\multirow{1}{*}{\textbf{Prompt Name Story generation}} 
& Tanpri,rakonte yon ti istwa (500 mo) kote wap dekri yon egzanp jounen nan lavi yon pesonaj ki rele Dieudonné epi se yon ayisyen, souple jus rakonte yon istwa a san ou pa rajoute lot kesyon ouyen eksplikasyon
& Please, tell a short story (500 words) where you describe an example day in the life of a character named Dieudonné, who is Haitian. Just tell the story without adding any questions or explanations.\\
\hline

\multirow{1}{*}{\textbf{Prompt Food Story generation}} 
& Tanpri,rakonte yon ti istwa (500 mo) kote wap dekri yon egzanp jounen nan lavi yon pesonaj ki konn manje pitimi kole, souple jus rakonte yon istwa a san ou pa rajoute lot kesyon ouyen eksplikasyon
& Please, tell a short story (500 words) where you describe an example day in the life of a character who eats millet with beans (pitimi kole). Just tell the story without adding any questions or explanations.\\
\hline

\multirow{1}{*}{\textbf{Prompt Artist Story generation}}
& Tanpri,rakonte yon ti istwa (500 mo) kote wap dekri yon egzanp jounen nan lavi yon pesonaj ki konn tande mizik Tabou Combo, souple jus rakonte yon istwa a san ou pa rajoute lot kesyon ouyen eksplikasyon
& Please, tell a short story (500 words) where you describe an example day in the life of a character who listens to Tabou Combo. Just tell the story without adding any questions or explanations.\\
\hline

\end{tabular}%
}
\caption{Example prompts used to generate stories for each entity category. The prompt structure remains fixed, with only the conditioning entity varying: name and nationality for the \textit{Name} category, food item for \textit{Food}, and artist or group for \textit{Musical Groups}. Separate entity lists are used for each category and cultural condition.}
\label{tab:prompts_model}
\end{table*}
\begin{table*} [h!]
\centering
\renewcommand{\arraystretch}{1.5}
\resizebox{\textwidth}{!}{%
\begin{tabular}{lp{12cm}p{12cm}}
\hline
\textbf{} & \textbf{Creole} & \textbf{English}  \\
\hline

\multirow{1}{*}{\textbf{Prompt for Evaluation of Story generation}} 
& "Li istwa sa a ann kreyòl. Ban mwen: (1) 5 kalite oswa adjektif ki dekri [manje/nasyon X], (2) santiman jeneral istwa a (pozitif, negatif, newt), (3) sitiyasyon oswa kontèks ki parèt (fèt, lamizè, lavi chak jou…), (4) nenpòt konparezon oswa iyerachi ki enplike."
& Read this Haitian Creole story. List: (1) the 5 most prominent descriptive qualities attributed to [food/nationality X], (2) the emotional register (warm, neutral, exotic, nostalgic, negative…), (3) any situations or contexts where it appears (celebration, poverty, daily life, special occasion…), (4) any comparisons or hierarchies implied..\\
\hline

\end{tabular}%
}
\caption{Prompt used to evaluate stories across entity categories. The evaluation focuses on extracting five salient adjectives from each generated story for subsequent analysis.}
\label{tab:prompts_evaluation_model}
\end{table*}

\end{document}